\documentclass[preprint,12pt]{elsarticle}
\usepackage{graphicx}
\usepackage{amssymb}
\usepackage{lineno}
\usepackage{hyperref}
\usepackage{amsmath}
\usepackage{algorithm} 
\usepackage{algpseudocode} 
\usepackage{enumerate}
\journal{preprint}
\usepackage{booktabs}
\usepackage{multirow} 
\usepackage{rotating}
\usepackage{pifont}
\usepackage{booktabs} 
\usepackage{tabularx} 
\usepackage{siunitx}  
\usepackage{ragged2e} 
\usepackage{array}    
\usepackage{subcaption}
\usepackage{tabularx} %
\usepackage{caption}  
\usepackage[table,xcdraw]{xcolor}
\usepackage{bbm}
\usepackage{dsfont}

\usepackage{lineno}
\usepackage{booktabs}
\usepackage{amsmath}

\begin{document}

\begin{frontmatter}


\title{VL-OrdinalFormer: Vision–Language Guided Ordinal Transformers for Interpretable Knee Osteoarthritis Grading}

\author{Zahid Ullah$^{1}$, Jihie Kim$^{1,*}$}

\address{%
$^{1}$ \quad Department of Computer Science and Artificial Intelligence, Dongguk University, Seoul 04620, Republic of Korea \\  } 

\begin{abstract}
Knee osteoarthritis (KOA) is a leading cause of disability worldwide, and accurate severity assessment using the Kellgren–Lawrence (KL) grading system is critical for clinical decision-making. However, radiographic distinctions between early disease stages, particularly KL1 and KL2, are subtle and frequently lead to inter-observer variability among radiologists. To address these challenges, we propose VL-OrdinalFormer, a vision–language-guided ordinal learning framework for fully automated KOA grading from knee radiographs. The proposed method combines a ViT-L/16 backbone with CORAL-based ordinal regression and a Contrastive Language–Image Pretraining (CLIP) driven semantic alignment module, allowing the model to incorporate clinically meaningful textual concepts related to joint space narrowing, osteophyte formation, and subchondral sclerosis. To improve robustness and mitigate overfitting, we employ stratified five-fold cross-validation, class-aware re-weighting to emphasize challenging intermediate grades, and test-time augmentation with global threshold optimization. Experiments conducted on the publicly available OAI kneeKL224 dataset demonstrate that VL-OrdinalFormer achieves state-of-the-art performance, outperforming CNN and ViT baselines in terms of macro F1-score and overall accuracy. Notably, the proposed framework yields substantial performance gains for KL1 and KL2 without compromising classification accuracy for mild or severe cases. In addition, interpretability analyses using Grad-CAM and CLIP similarity maps confirm that the model consistently attends to clinically relevant anatomical regions. These results highlight the potential of vision–language-aligned ordinal transformers as reliable and interpretable tools for KOA grading and disease progression assessment in routine radiological practice. Code is available at: \url{https://github.com/Zahid672/kneeKL224}
\end{abstract}


\end{frontmatter}

\section{Introduction}
\label{intro}
\begin{figure*}[!ht]
\centering
\includegraphics[width=1\textwidth]{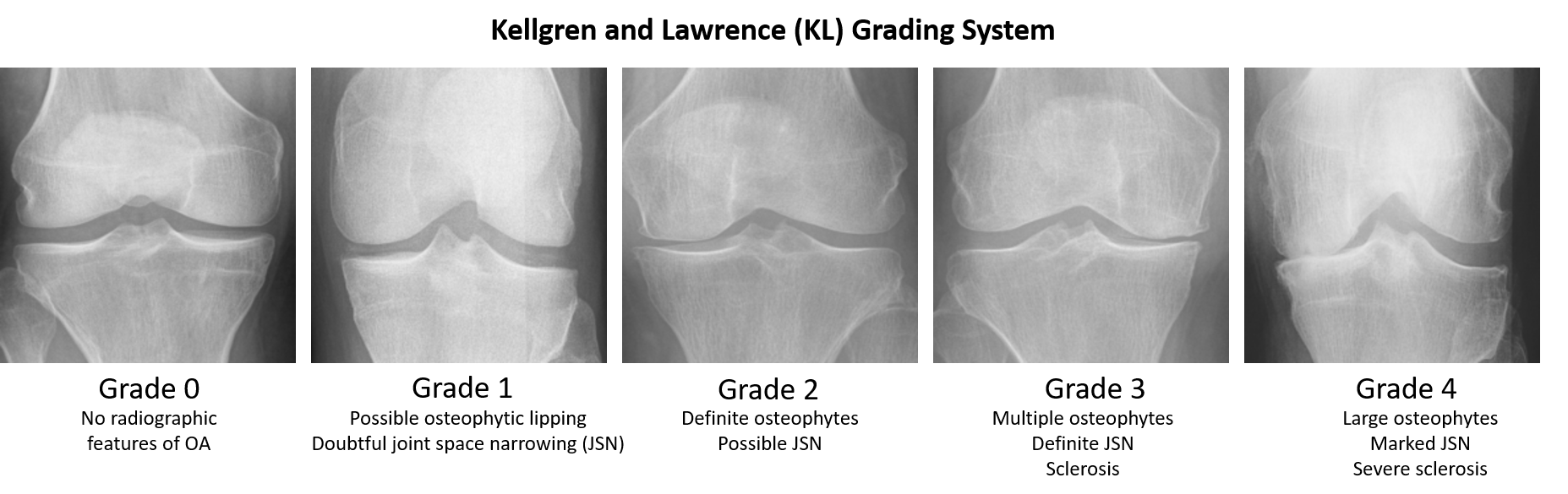}
\caption{Knee joint samples of all KL grades and their corresponding criterion.}
\label{kl_grades}
\end{figure*}

Knee osteoarthritis (KOA) is one of the most common degenerative joint diseases worldwide and a leading cause of disability in older adults \cite{losina2013disease}. Early identification of structural deterioration is critical for timely intervention, treatment planning, and preventing long-term disability. Radiographic assessment using the Kellgren–Lawrence (KL) grading system remains the clinical standard for KOA severity evaluation \cite{kellgren1957radiological}. In clinical environments, accurate KL grading can help orthopedists determine disease progression, select weight-bearing therapies, and avoid unnecessary surgical procedures. However, manual grading requires expert radiologists and is inherently subjective, leading to significant inter-observer variability. Each grade sample and criterion is illustrated in Fig. \ref{kl_grades}.

Recent advances in deep learning have enabled automated KL grade classification from knee radiographs, demonstrating performance comparable to expert readers in some settings \cite{farooq2023dc,tariq2025review}. Despite this progress, several challenges remain unresolved. First, the visual boundary between early disease stages, particularly KL-1 and KL-2, is subtle and characterized by marginal radiographic findings such as doubtful osteophytes or mild joint space narrowing. This ambiguity frequently leads to confusion between adjacent grades, limiting the clinical reliability of automated systems. Second, most existing approaches \cite{kwon2020machine,maqsood2025knee} rely solely on image-based cues and lack explicit incorporation of clinical semantics. While such models often perform well for advanced KOA stages (KL-3 and KL-4), where structural deformities are pronounced, they struggle to capture the nuanced pathological changes present in borderline cases. Third, achieving balanced sensitivity and specificity across all grades remains challenging due to class imbalance and heterogeneity in radiographic appearance.

For automated KOA grading systems to be clinically deployable, they must not only achieve high accuracy but also reason in a medically meaningful manner and provide transparent explanations for their predictions. This requirement motivates the integration of vision–language alignment, where radiological knowledge is explicitly encoded using textual descriptions and leveraged as an additional supervisory signal. Vision–Language Models (VLMs) \cite{bordes2024introduction,sohail2025knee}, such as Contrastive Language–Image Pretraining (CLIP) \cite{radford2021learning}, learn joint representations of images and text, enabling alignment between visual patterns and semantic concepts. In the context of KOA grading, aligning model predictions with clinically grounded textual descriptions, such as “definite osteophytes,” “moderate joint space narrowing,” or “subchondral sclerosis,” can guide the learning process toward disease-relevant attributes and improve interpretability. 

In this work, we present a vision–language-guided ordinal learning framework for automated KOA severity assessment from knee radiographs. The proposed approach integrates a Vision Transformer (ViT) backbone with CORAL-based ordinal regression to explicitly model the ordered nature of KL grades. In addition, CLIP-based semantic supervision is employed to align visual features with medically meaningful textual descriptions of KOA pathology. This design enables the model to capture subtle radiographic distinctions between adjacent grades while maintaining strong performance for advanced disease stages. Overall, this study demonstrates that integrating domain-specific knowledge through vision–language alignment significantly improves the reliability, interpretability, and clinical relevance of automated KOA severity grading, highlighting its potential for supporting routine radiological assessment and longitudinal disease monitoring. The main contributions of this study are summarized as follows:

\begin{itemize}
\item We propose a vision–language-guided ViT-CORAL framework that incorporates clinical text descriptions of KL grades into the training process, encouraging attention to radiologically relevant features.
\item We improve ordinal KOA classification by employing a CORAL-based decision head tailored to ordered severity levels, reducing misclassification between adjacent grades such as KL-1 and KL-2.
\item We introduce a text-contrastive supervision and feature distillation module that leverages CLIP embeddings to align visual representations with semantic knowledge related to osteophytes, joint space narrowing, and subchondral sclerosis.
\item We adopt an ensemble inference strategy with test-time augmentation (TTA) and global threshold optimization, achieving state-of-the-art performance on the OAI knee dataset while preserving balanced accuracy across KL categories.
\item We enhance clinical interpretability through text-prompt similarity analysis and visual explanation techniques, providing insight into the radiographic cues driving model predictions.
\end{itemize}

The remaining paper is organized as follows: Section \ref{rw} presents a comprehensive review of related work to contextualize our research. Further, we discuss Dataset and preprocessing in Section \ref{dataset}. Next, in Section \ref{pm} we detail our proposed methodology, including its overview, and main architecture. We then describe our experimental setup, including baseline comparison, ablation studies in section \ref{experimental}. Section \ref{inter} presents the interpretability and analysis, whereas Section \ref{discussion} consists of the discussion. Finally, in section \ref{con}, we present the conclusion.

\section{Related Work}\label{rw}
\subsection{Radiographic KOA grading}
The KL system is the standard radiographic scale for assessing KOA severity. It grades structural changes from 0 (no radiographic features of OA) to 4 (severe joint space loss and bone deformity), based on the presence and extent of osteophytes, joint space narrowing, subchondral sclerosis, and bony deformity. KL grading is routinely used in clinical practice and large cohort studies such as the Osteoarthritis Initiative (OAI), since it provides a simple, discrete summary of disease status that can be used for diagnosis, prognosis, and treatment planning \cite{pi2023ensemble}. However, KL grading is inherently subjective. Multiple studies have shown considerable inter and intra-observer variability, particularly in early stages such as KL-1 and KL-2, where radiographic changes are subtle and sometimes overlap with normal anatomical variation \cite{zhao2025value}. This variability affects patient stratification in clinical trials and can obscure real treatment effects. It also motivates the development of automated or decision-support systems that can produce more consistent assessments.

Traditional computer-aided diagnosis approaches \cite{yunus2022recognition,sharma2020conventional} for KOA used hand-engineered features, such as joint space width, shape descriptors, or texture features extracted from defined regions of interest. These were often combined with classical machine learning classifiers, but typically required careful preprocessing, manual tuning, and did not generalize well across datasets or imaging protocols. Recent work has shifted toward end-to-end deep learning methods that operate directly on X-ray images, aiming for fully automatic KL grading. Several recent papers have shown that deep learning models can reach performance comparable to experienced radiologists on KL grading tasks. For example, Vaattovaara et al.  \cite{vaattovaara2025kellgren} developed a deep learning model for KL grading that achieved diagnostic accuracy comparable to highly experienced human readers in an external validation cohort.
Other studies  \cite{pan2024automatic} have explored multi-view radiographs, hierarchical classification schemes, and integration of prior anatomical knowledge to improve robustness and clinical relevance. Despite these advances, early KOA detection (KL-1 and KL-2) remains a key weakness across most existing systems, with performance consistently higher for KL-3 and KL-4 than for mild disease \cite{zhao2025value}. 

\subsection{CNN-based KOA classification} 
Deep CNNs were the first family of models to substantially improve automatic KL grading. A landmark work by Chen et al.  \cite{chen2019fully} proposed a fully automatic KOA severity grading pipeline based on deep neural networks. Their system first used a customized YOLOv2 detector to localize knee joints in full radiographs and then applied a fine-tuned VGG-19 classifier with a novel ordinal loss on the OAI dataset. The proposed ordinal formulation explicitly encoded the ordered structure of KL grades in the loss function, and achieved an accuracy of about 69.7\% and a mean absolute error of 0.344 on the KL grading task, outperforming earlier CNN baselines. This work is frequently used as a reference baseline for subsequent KOA grading studies that adopt the same dataset and evaluation splits.

Later studies extended CNN-based KOA grading in several directions. Swiecicki et al. \cite{swiecicki2021deep} developed a fully automatic deep learning algorithm for KL grading and evaluated it on multi-institutional data, reporting quadratic kappa scores that approached the agreement between human readers.
 Pan et al. \cite{pan2024automatic} proposed a hierarchical classification framework that separates the detection of OA from the estimation of its severity, which aims to reflect radiologists’ reasoning and improves performance on intermediate grades.
 Other works compared a variety of CNN architectures, including VGG-16/19, ResNet-101, and EfficientNet variants, to identify the most suitable backbone for KL grading, again highlighting that performance gains are often largest for advanced OA and more modest for early stages \cite{chen2019fully}.

Ensemble CNN approaches have also been explored. Pi et al. \cite{pi2023ensemble} used ensemble deep-learning networks to improve the robustness of KL classification and showed that aggregating multiple models reduced prediction variance and increased diagnostic performance, especially for KL-3 and KL-4. Nasef et al. \cite{nasef2024deep} evaluated different deep learning and classical machine learning models for KL grading on multiple public datasets and reported that, while deep models clearly outperform traditional approaches, they still face challenges related to class imbalance and dataset heterogeneity.

Overall, CNN-based KOA classification methods have demonstrated that fully automatic KL grading is feasible and can reach strong performance, particularly for moderate and severe OA. However, they are limited by relatively local receptive fields, reliance on image-only supervision, and a lack of mechanisms to explicitly incorporate clinical definitions or ordinal structure beyond the loss design.

\subsection{Vision Transformers in medical imaging}

ViTs \cite{dosovitskiy2020image} have recently emerged as competitive alternatives to CNNs for visual representation learning. Unlike CNNs \cite{krizhevsky2012imagenet}, which build features through progressively larger convolutional receptive fields, ViTs operate on image patches using self-attention, which allows them to model long-range dependencies and global context more naturally. This property is especially useful in medical imaging tasks where clinically relevant patterns may span larger anatomical regions or multiple structures.

Several surveys provide comprehensive overviews of Transformer architectures in medical imaging \cite{shamshad2023transformers,kumar2024applications,khan2025recent}, covering tasks such as classification, segmentation, detection, reconstruction, and report generation. Shamshad et al. \cite{shamshad2023transformers} summarized more than 125 Transformer-based methods and outlined how pure ViTs and hybrid CNN–Transformer architectures have been applied to diverse modalities including X-ray, CT, MRI, ultrasound, and histopathology.
 More recent reviews have specifically focused on ViTs in classification settings, highlighting their advantages in capturing global context and their limitations when data is scarce or highly imbalanced \cite{khan2025recent}. 
In biomedical image classification, several works have demonstrated that ViTs can match or surpass CNN performance when combined with appropriate pretraining and data augmentation. Halder et al.,  \cite{halder2024implementing} systematically evaluated ViTs on MedMNIST classification tasks and showed that they can provide competitive performance with suitable optimization and regularization strategies.
 Transformer-based models have also been successfully used in segmentation, where architectures like TransUNet, Swin-UNet, and hybrid CNN–ViT designs capture both local structure and global anatomy \cite{shamshad2023transformers}.

Despite these advances, ViTs in KOA grading are still relatively underexplored compared to CNNs. The potential benefits are clear; for instance, ViTs can aggregate information across the whole knee joint and better capture global patterns such as alignment, multi-compartment joint space narrowing, and diffuse osteophyte distribution. However, they also require careful handling of dataset size, class imbalance, and regularization to avoid overfitting. In this context, combining ViTs with ordinal losses and ensemble inference, as done in this work, is a natural extension of earlier CNN-based KOA grading systems.

\subsection{Vision-Language Models in Clinical AI}

Vision-Language Models (VLMs) \cite{bordes2024introduction} extend the idea of representation learning by jointly modeling images and text in a shared embedding space. CLIP \cite{radford2021learning} is a prominent example that is trained on large collections of image-text pairs \cite{schuhmann2021laion400m} to align visual content with natural language descriptions. This alignment enables zero-shot classification, cross-modal retrieval, and text-guided interpretation without task-specific retraining. In the clinical domain, this capability is particularly appealing because radiology practice routinely integrates images with textual reports, and many diagnostic criteria are defined using structured medical terminology.

Recent years have seen rapid growth in adapting CLIP-like models to medical imaging. MedCLIP \cite{wang2022medclip}, for instance, learns from unpaired chest X-ray images and radiology reports and has demonstrated strong performance on both zero-shot and supervised classification tasks. Zhao et al.\ \cite{zhao2025clip} conducted a comprehensive survey of CLIP in medical imaging, summarizing pretraining strategies, datasets, and downstream applications ranging from classification to dense prediction and multimodal retrieval. Other models such as PMC-CLIP \cite{lin2023pmc} and RadCLIP \cite{ruckert2024rocov2} leverage large biomedical literature corpora or radiology text archives to develop domain-adapted VLMs capable of disease classification, report generation, and visual question answering.

A consistent theme in these studies is that VLMs \cite{bordes2024introduction} often improve interpretability and robustness relative to image-only models. Because VLMs operate in a joint vision–language space, predictions can be grounded in human-readable concepts, enabling more transparent error analysis and semantic reasoning. Recent work has shown that text prompts capturing pathology descriptions can guide models toward clinically relevant regions even when annotated data are limited, enabling weakly supervised classification and zero-shot anomaly detection \cite{zhao2025clip}.

Despite this progress, the use of VLMs for ordinal, disease-staging problems such as KL-based KOA grading remains largely unexplored. Most existing CLIP-based medical applications target binary, multilabel, or standard multiclass classification settings, where classes are assumed to be independent and unordered. In contrast, KOA severity follows a strict radiological progression, and neighboring grades (e.g., KL1 vs.\ KL2) differ only by subtle morphological cues. This creates a fundamentally different learning problem that benefits from ordinal regression rather than standard classification. Furthermore, grade-specific textual criteria such as “definite osteophytes with possible joint space narrowing’’ for KL2 provide rich semantic information that has not been incorporated into prior CLIP-based medical models.

Existing VLM frameworks such as MedCLIP, PMC-CLIP, and RadCLIP primarily address classification or report-generation tasks and do not incorporate ordinal regression, threshold-based decision functions, or disease staging supervision. Moreover, these models focus on natural alignment between images and free-text radiology reports, whereas the present work leverages structured, grade-specific radiological definitions as explicit text prompts. Our method is therefore distinct in two key aspects: (1) it integrates VLM-derived semantic supervision into a ViT-based CORAL ordinal regression framework tailored to KL grading, and (2) it uses CLIP text embeddings as auxiliary targets to improve fine-grained discrimination between adjacent disease stages. To our knowledge, this constitutes the first integration of CLIP-style text supervision within an ordinal KOA severity grading pipeline.

\section{Dataset and Preprocessing}\label{dataset}

\subsection{OAI Knee Radiograph Dataset}

This study utilizes knee radiographs from the publicly available OAI cohort \cite{nevitt2006osteoarthritis}, which includes longitudinal imaging and clinical information collected from nearly 4,796 individuals. The posterior-anterior (PA) fixed-flexion X-ray view was selected, as it is widely used for clinical assessment of structural degeneration in KOA. Each radiograph consists of two knee joints, which were cropped into individual joint images for automated KL grading. All images were originally stored in grayscale format and converted to three channels for compatibility with ImageNet-pretrained deep learning architectures. 

\subsection{KL Grading Categories and Class Imbalance}

The KL grading system assigns KOA severity across five ordered categories, such as, KL-0 represent no radiographic features of OA, KL-1 shows doubtful osteophytes, questionable joint space narrowing, KL-2 denotes definite osteophytes, and possible narrowing, KL-3 shows moderate joint space narrowing, sclerosis, and possible deformity, and KL-4 provides severe narrowing with large osteophytes and bone deformity.

\begin{figure*}[!ht]
\centering
\includegraphics[width=0.7\textwidth]{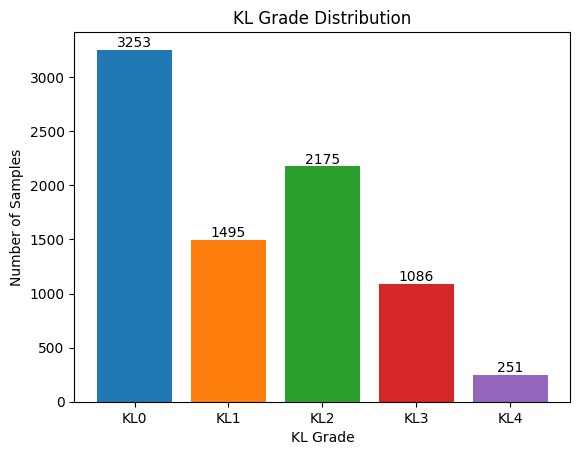}
\caption{KL grade distribution in the OAI knee radiograph dataset.}
\label{histogram}
\end{figure*}

As shown in Fig. \ref{histogram}, the KOA dataset exhibits a strong class imbalance, with KL0 and KL2 being the most common grades, while KL3 and especially KL4 appear far less frequently. This distribution reflects typical patterns in population-level radiographic datasets but presents challenges for deep learning models, which tend to underperform on underrepresented classes. Our approach explicitly accounts for this imbalance through class-weighted CORAL loss, ensemble inference, and CLIP-based semantic regularization, helping the model better discriminate between adjacent grades despite uneven sample availability. 

\begin{figure*}[t]
\centering
\includegraphics[width=0.8\textwidth]{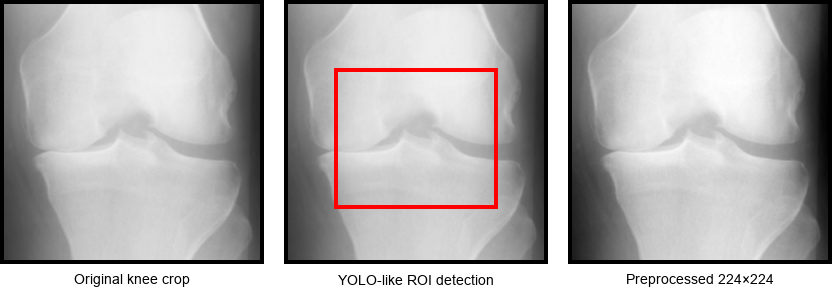}
\caption{Illustration of the preprocessing pipeline.
From left to right: (a) original knee crop, (b) YOLO-like localization of the
diagnostic region of interest, and (c) final preprocessed 224$\times$224 patch
used as input to the ViT-CORAL model.}
\label{fig:knee_yolo_pipeline}
\end{figure*}

\subsection{YOLO-based Joint Cropping}

To eliminate irrelevant background regions in full knee radiographs, we employ a deep-learning-based knee localization approach using a YOLOv2 detector trained to identify left and right knee joints. Detected joints are cropped with approximately 1.3$\times$ bounding box expansion to retain anatomical context, including medial/lateral compartments and periarticular structures necessary for severity assessment. This preprocessing step yields standardized region-of-interest patches containing only diagnostic image content.

\subsection{Train/Validation/Test Splits}

To ensure an unbiased performance evaluation, a stratified 5-fold cross-validation protocol is adopted. In each fold, three folds are used for model training. One fold serves as validation for model selection, and the remaining fold is used for independent testing. All splits are created at the subject level to prevent information leakage, ensuring that knee joints extracted from the same patient do not appear across training and testing sets. Final performance is reported using the ensemble and best-fold results aggregated across all folds.

\subsection{Image Preprocessing and Data Augmentation}

All knee joint images are resized to 224$\times$224 pixels and normalized using ImageNet mean and standard deviation statistics. To improve robustness against limited high-grade training samples, data augmentation is applied during training, consisting of Random resized cropping (scale: 0.8-1.0), random horizontal flipping, and random rotation ($\pm10^{\circ}$). Center cropping and deterministic normalization are used during validation and testing to ensure consistent and reproducible evaluation conditions. Table~\ref{tab:prep_summary} summarizes the major challenges addressed during preprocessing and their corresponding mitigation strategies.

\begin{table}[h]
\centering
\caption{Overview of preprocessing strategies and their motivation.}
\label{tab:prep_summary}
\begin{tabular}{p{3.0cm} p{5.5cm} p{4.5cm}}
\toprule
\textbf{Challenge} & \textbf{Method Applied} & \textbf{Impact} \\
\midrule
High background clutter &
YOLO-based knee joint localization &
Focuses on diagnostic anatomy \\
Class imbalance &
Class-weighted ordinal modeling &
Improves recognition of mild OA \\
Data variability &
On-the-fly augmentation &
Enhanced generalization \\
Overfitting risk &
5-fold cross-validation &
Reliable performance reporting \\
\bottomrule
\end{tabular}
\end{table}

\section{Methodology}\label{pm}
\subsection{Overview}

\begin{figure*}[!ht]
\centering
\includegraphics[width=0.6\textwidth]{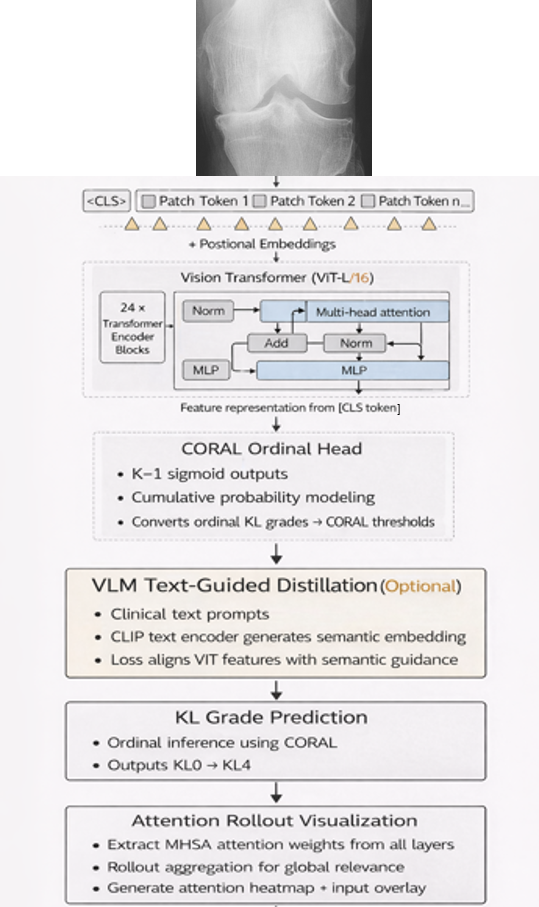}
\caption{Schematic overview of the proposed vision-language-guided framework for automatic KOA severity prediction. The input knee X-ray image is processed through a ViT-L/16 backbone with 16×16 patch embedding. Feature representations from the [CLS] token are passed through a CORAL ordinal regression head to predict KL grades (0–4). An optional VLM distillation module aligns visual features with clinical text embeddings from CLIP. Finally, attention rollout visualizations highlight the anatomical regions influencing the model’s decisions.}
\label{schematic}
\end{figure*}

This section presents the methodological framework of the proposed VL-OrdinalFormer as illustrated in Fig. \ref{schematic}. We begin by establishing baseline comparisons using a VGG19 and ViT backbone trained with a conventional multi-class classification objective. Next, we introduce an ordinal modeling formulation based on Vision Transformers with CORAL (ViT-CORAL), which explicitly accounts for the ordered nature of KL grades. We then incorporate a Vision–Language Distillation Module (VLM) that leverages CLIP-based semantic supervision to align visual representations with clinically meaningful textual descriptions. Finally, we describe the overall training strategy, including the loss formulation, class-aware re-weighting, and optimization details, and conclude with an ensemble-based inference scheme combined with TTA and global threshold tuning.

\subsection{Baseline Methods}

\subsubsection{VGG19 Baseline}
We replicate the widely adopted experimental protocol of Chen et al. \cite{chen2019fully}, which employs a fixed preprocessing pipeline consisting of knee joint localization using YOLOv2, a $1.3\times$ region-of-interest expansion, and resizing the cropped knee region to a resolution of $224\times224$ pixels. A VGG19 network pretrained on ImageNet is fine-tuned for five-class KL grade classification using a standard softmax output layer and categorical cross-entropy loss. As in the original baseline, no explicit ordinal structure is imposed on the output space, and the classifier treats KL grades as independent categories despite their natural ordering from KL0 to KL4. This baseline serves as a direct reference point for evaluating the effectiveness of ordinal modeling and vision–language integration introduced in our proposed approach.

\subsubsection{Standard Multi-class Cross-entropy}
In addition to the CNN baseline, we evaluate a Vision Transformer (ViT–B/16) trained using a conventional five-class softmax classifier with categorical cross-entropy loss. This model similarly ignores ordinal relationships among KL grades and predicts each class independently. By comparing this baseline with ordinal and vision–language-enhanced variants, we isolate the benefits contributed by CORAL-based ordinal regression and semantic supervision.

\subsection{Proposed ViT-CORAL Model}

\subsubsection{Transformer Backbone}
We adopt Vision Transformer architectures (ViT–B/16 and ViT–L/16) pretrained on ImageNet-1K as the visual backbone. Input knee radiographs are partitioned into fixed-size patches and processed through multiple self-attention layers, enabling the model to capture long-range contextual dependencies across the joint region. The final [CLS] token embedding is used as a global image representation and passed to subsequent prediction heads.

\subsubsection{Ordinal Regression with CORAL}
To explicitly model the ordered nature of KL grades, we employ the COnsistent RAnk Logits (CORAL) \cite{caoarank} formulation for ordinal regression. Instead of predicting $K$ independent class probabilities ($K=5$ for KL0–KL4), the model outputs $K-1$ binary decision thresholds:

\begin{equation}
P(y > k) = \sigma(z_k), \quad k = 0, 1, \dots, K-2,
\end{equation}

where $z_k$ denotes the learned logits and $\sigma(\cdot)$ is the sigmoid function. During inference, the predicted KL grade is obtained by counting the number of thresholds for which $P(y > k) \ge \tau$, where $\tau$ is a tunable global decision threshold. This formulation offers several advantages, such that it enforces monotonicity among decision boundaries, secondly, it embeds ordinal structure directly into the learning objective, and lastly, it improves separability between clinically ambiguous adjacent grades, particularly KL1 and KL2.

\subsubsection{Class-aware Weighting Strategy}
The OAI dataset exhibits notable class imbalance, and mid-grade categories (KL1–KL2) are especially prone to confusion due to subtle morphological differences. To mitigate this issue, we introduce a class-aware weighting strategy defined as:

\begin{equation}
\mathbf{w} = [1.0, \alpha, \alpha, 1.0, 1.0], \quad \alpha > 1,
\end{equation}

which places greater emphasis on intermediate grades during training. Empirically, setting $\alpha = 1.5$ yielded the most stable optimization behavior and resulted in substantial recall improvements for KL1 and KL2 without degrading performance for mild or severe grades.

\subsection{Vision-Language Distillation Module}
To enhance semantic understanding and clinical relevance, we integrate a lightweight VLM based on CLIP embeddings. Each KL grade is associated with a clinically grounded textual description, for example (i) no radiographic osteoarthritis (KL0), (ii) doubtful joint space narrowing” (KL1), (iii) definite osteophytes, possible narrowing” (KL2), (iii) multiple osteophytes, definite narrowing (KL3), and severe narrowing and subchondral sclerosis (KL4).

Textual descriptions are encoded using the CLIP text transformer, while visual features are extracted from the ViT–CORAL backbone. A contrastive alignment loss is then applied:
\begin{equation}
\mathcal{L}_{\text{CLIP}} =
-\log \frac{\exp\!\left(\langle f_{\text{img}}, f_{\text{text}} \rangle / \tau \right)}
{\sum_{j} \exp\!\left(\langle f_{\text{img}}, f_{\text{text}_j} \rangle / \tau \right)},
\end{equation}

where $f_{\text{img}}$ and $f_{\text{text}}$ denote normalized image and text embeddings, respectively, and $\tau$ is a temperature parameter. This objective encourages semantically consistent alignment between visual representations and medically meaningful textual concepts, leading to improved generalization, enhanced interpretability, and better discrimination of subtle KOA severity levels.

\subsection{Training Strategy}

\subsubsection{Loss Function}
The total learning objective combines:

\begin{equation}
\mathcal{L} =
\mathcal{L}_{\text{CORAL}}
+ \lambda \mathcal{L}_{\text{CLIP}}
+ \mu \mathcal{L}_{\text{reg}},
\end{equation}

where $\mathcal{L}_{\text{CORAL}}$ denotes the weighted binary cross-entropy loss applied to the ordinal thresholds, $\mathcal{L}_{\text{CLIP}}$ enforces vision–language semantic alignment, and $\mathcal{L}_{\text{reg}}$ represents $L_2$ regularization to promote stable optimization. The weighting coefficients $\lambda$ and $\mu$ are selected empirically based on validation performance.

\subsubsection{Optimization and Regularization}
All models are optimized using the AdamW optimizer with an initial learning rate of $3\times10^{-5}$ and a cosine annealing learning rate schedule. To enhance generalization, we apply standard data augmentation techniques, including normalized resizing, random horizontal flipping, and random rotations within $\pm10^\circ$. Early stopping with a patience of 10 epochs, monitored on validation accuracy, is employed to prevent overfitting.

\subsection{Ensemble Inference and Threshold Tuning}

To improve robustness and reduce variance, we adopt a five-fold ensemble inference strategy. Models trained on different validation splits are combined at test time, and TTA is applied by averaging prediction logits across multiple spatial transformations for each fold. The final prediction is obtained via majority voting across the five folds. To further refine ordinal decision boundaries, the global threshold $\tau$ used for CORAL decoding is tuned independently on each fold’s validation set by maximizing the macro F1-score. This post-training threshold optimization improves discrimination between adjacent grades, particularly in mid-grade transitions, and yields enhanced test performance without requiring additional model retraining.

\section{Evaluation Methodology}

\subsection{5-Fold Cross-Validation and Ensemble Inference}

We employ a stratified five-fold cross-validation strategy to ensure that the distribution of KL grades is preserved across all data splits. For each fold, the dataset is partitioned into three subsets:
(i) a training set,
(ii) a validation set used for model selection and threshold optimization, and
(iii) a held-out test set reserved for final performance reporting.

A separate ViT–CORAL model is trained for each fold, resulting in five independently optimized models. During inference, predictions from all models are aggregated through logit averaging:
\begin{equation}
\hat{z} = \frac{1}{5} \sum_{i=1}^{5} z^{(i)},
\end{equation}

where $z^{(i)}$ denotes the output logits of the model trained on the $i$-th fold. Final class labels are obtained by applying the optimized ordinal thresholding scheme described in Section~\ref{subsec:tau}. This ensemble strategy reduces sensitivity to variations in training splits and improves prediction stability, leading to stronger generalization across diverse clinical sub-populations.

\subsection{Test-Time Augmentation}
To further enhance robustness, TTA is applied during inference. For each test radiograph $x$, a set of augmented variants ${x_j}_{j=1}^{N}$ is generated using the following transformations, original view, horizontal flip, $+10^\circ$ rotation, $-10^\circ$ rotation, with $N = 4$ in all experiments. Model logits are averaged across all augmented predictions prior to ordinal thresholding. This strategy mitigates prediction variance caused by minor anatomical misalignments, detector localization noise, and subtle pose variations.

\subsection{Threshold Optimization for Ordinal Boundaries}
\label{subsec:tau}

Rather than adopting a fixed decision threshold ($\tau = 0.5$) for CORAL decoding, we optimize the threshold value on the validation set to maximize the macro F1-score:
\begin{equation}
\tau^{*} = \arg\max_{\tau \in [0,1]} \, \text{F1}_{\text{macro}}(\tau).
\end{equation}

This adaptive calibration sharpens discrimination between adjacent ordinal grades and reduces bias toward majority classes, particularly for underrepresented mid-grade categories such as KL1 and KL2. The optimized threshold $\tau^{*}$ is subsequently applied consistently to all test samples without requiring model retraining.

\subsection{Evaluation Metrics}

Given the ordinal nature and class imbalance inherent in KL grading, we report a comprehensive set of performance metrics to provide a balanced and clinically meaningful evaluation.

\subsubsection{Accuracy}
Overall classification accuracy is defined as the proportion of correctly predicted samples:
\begin{equation}
\text{Accuracy} = \frac{1}{N}\sum_{i=1}^{N} \mathbf{1}\{\hat{y}_i = y_i\}.
\end{equation}

\subsubsection{Macro- and Weighted-Averaged Precision, Recall, and F1-score}
To account for class imbalance, both macro- and weighted-averaged metrics are reported. Macro metrics assign equal importance to each class, whereas weighted metrics reflect true class frequencies:
\begin{equation}
\text{F1}_{\text{macro}} = \frac{1}{K}\sum_{c=1}^{K} \text{F1}_{c},
\qquad
\text{F1}_{\text{weighted}} =
\sum_{c=1}^{K} \frac{N_{c}}{N}\,\text{F1}_{c}.
\end{equation}


\subsubsection{AUROC (One-vs-Rest)}
The area under the receiver operating characteristic curve (AUROC) is computed for each class using a one-vs-rest strategy and averaged across all classes:
\begin{equation}
\text{AUROC}_{\text{macro}} =
\frac{1}{K}\sum_{c=1}^{K} \text{AUROC}_{c}.
\end{equation}

This metric evaluates class separability independent of class imbalance.


\subsection{Confusion Matrix Analysis}

\begin{figure*}[!ht]
\centering
\includegraphics[width=0.7\textwidth]{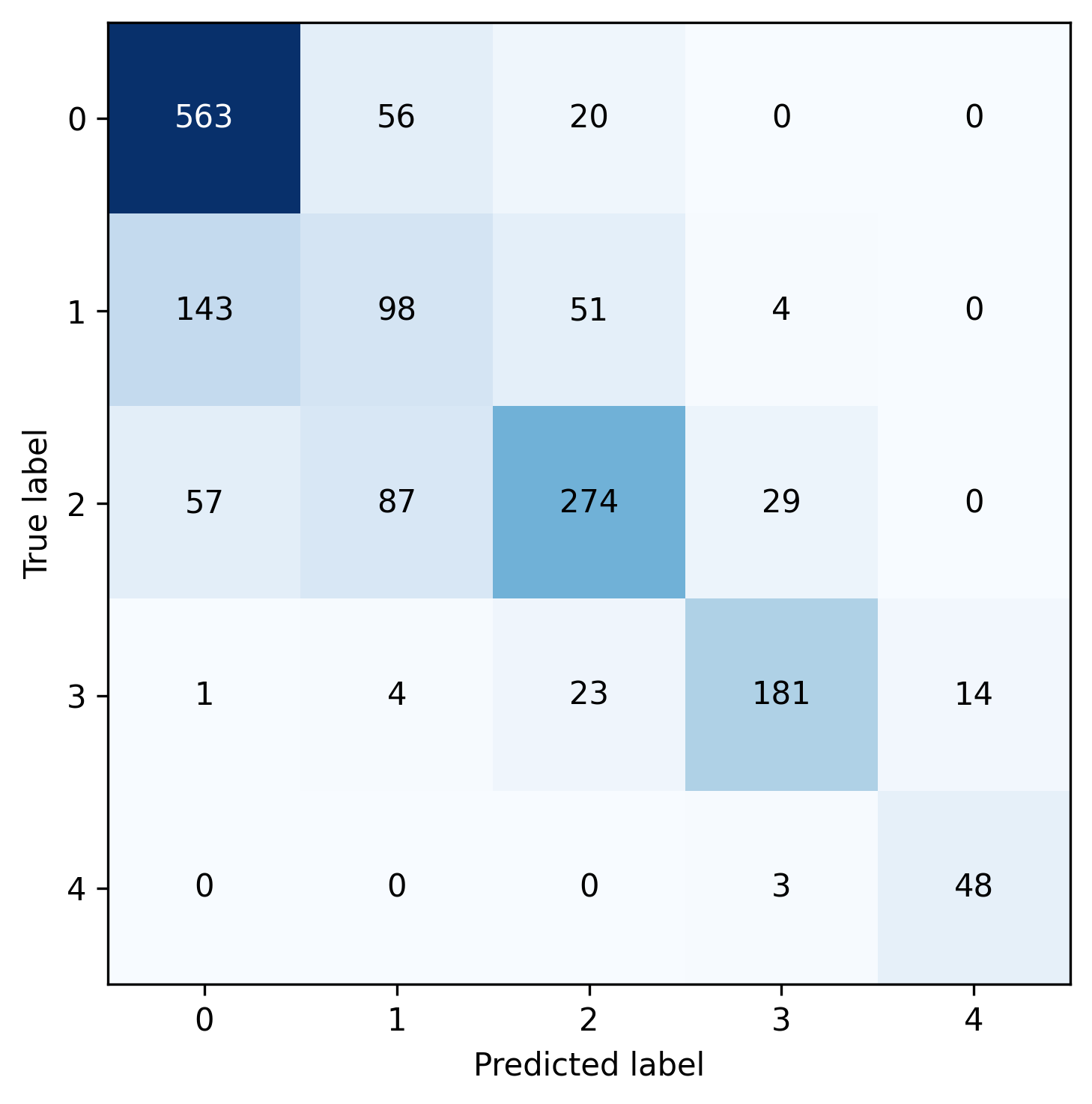}
\caption{Confusion matrix.}
\label{fig:confusion_matrix}
\end{figure*}

Confusion matrices are used to visualize pairwise misclassification patterns between KL grades. As illustrated in Fig.~\ref{fig:confusion_matrix}, most errors occur between adjacent ordinal categories, reflecting the intrinsic difficulty of distinguishing subtle transitions such as KL1–KL2 and KL2–KL3. Notably, the proposed ViT–CORAL with vision–language guidance and ensemble inference substantially reduces severe misclassifications (e.g., KL4 $\rightarrow$ KL0), preserves diagnostic sensitivity for early-stage KOA,
and produces error patterns consistent with clinically reasonable uncertainty. These findings reinforce the interpretability, robustness, and clinical relevance of the proposed framework.

\section{Experimental Results}\label{experimental}

This section presents a comprehensive evaluation of the proposed framework under multiple experimental settings, including a reproduced VGG19 baseline following Chen et al., \cite{chen2019fully}, the proposed ViT–CORAL architecture, incremental improvements introduced by the VLM, ensemble inference with TTA and ordinal threshold tuning, and class-wise performance analysis across KL grades. All final results are reported on the official held-out OAI test set, which consists of 1,656 knee radiographs spanning five KL grades (KL0–KL4). Table \ref{comparison} presents a comparative analysis of our proposed model with previous state-of-the-art models.

\begin{table}[!ht]
\centering
\caption{Comparison of baseline models with the proposed method on KL-grade classification. The proposed model shows superior results across all reported metrics.}
\label{comparison}
\resizebox{\textwidth}{!}{
\begin{tabular}{lcccccc}
\hline
Model &
\begin{tabular}[c]{@{}c@{}}Accuracy\\ (\%)\end{tabular} &
\begin{tabular}[c]{@{}c@{}}Precision$_{macro}$\\ (\%)\end{tabular} &
\begin{tabular}[c]{@{}c@{}}Recall$_{macro}$\\ (\%)\end{tabular} &
\begin{tabular}[c]{@{}c@{}}F1$_{macro}$\\ (\%)\end{tabular} &
\begin{tabular}[c]{@{}c@{}}Specificity$_{macro}$\\ (\%)\end{tabular} &
\begin{tabular}[c]{@{}c@{}}AUROC$_{macro}$\\ (\%)\end{tabular} \\
\hline
Deep Siamese CNN \cite{tiulpin2018automatic} & 66.71 & -- & -- & -- & -- & -- \\
ResNet101 \cite{mohammed2023knee} & 69.00 & 67.00 & -- & 65.00 & -- & -- \\
MobileNetV2 \cite{mohammed2023knee} & 67.00 & 69.00 & -- & 67.00 & -- & -- \\
VGG-19 \cite{chen2019fully} & 69.60 & -- & -- & -- & -- & -- \\
DenseNet121 \cite{chen2019fully} & 68.20 & -- & -- & -- & -- & -- \\
InceptionV3 \cite{chen2019fully} & 68.40 & -- & -- & -- & -- & -- \\
\textbf{Proposed} & \textbf{70.29} & \textbf{69.80} & \textbf{71.56} & \textbf{70.19} & \textbf{92.18} & \textbf{81.61} \\
\hline
\end{tabular}
}
\end{table}

\subsection{Baseline Comparison with Previous Work}

Table~\ref{tab:baseline_comparison} presents a direct comparison between reproduced CNN- and transformer-based baselines and the proposed ViT–CORAL framework. In addition to the VGG19 ordinal baseline, we also reproduce a plain ViT trained with cross-entropy loss, which achieves an accuracy of 68.59\% and a macro F1-score of 69.42\%. This result confirms that global self-attention alone provides a noticeable improvement over conventional CNN architectures, but remains limited when disease severity is treated as a set of independent classes. In contrast, the proposed ViT–CORAL with ensemble inference and TTA achieves 70.29\% accuracy and a 70.19\% macro F1-score, corresponding to a +1.7\% absolute accuracy gain and a +0.8\% macro F1 improvement over the reproduced ViT baseline, and a +5.5\% absolute accuracy improvement over the reproduced VGG19 ordinal model. These results demonstrate that explicitly modeling the ordinal structure of KL grades provides complementary benefits beyond those obtained from transformer architectures alone.

\begin{table}[!ht]
\centering
\resizebox{\textwidth}{!}{
\begin{tabular}{ccccc}
\hline
\textbf{Method} & 
\textbf{Architecture} & 
\textbf{Loss} & 
\textbf{\begin{tabular}[c]{@{}c@{}}Accuracy\\ (\%)\end{tabular}} & 
\textbf{\begin{tabular}[c]{@{}c@{}}Macro-F1\\ (\%)\end{tabular}} \\
\hline

Chen et al. \cite{chen2019fully} & VGG19 & Adjusted Ordinal & 69.7 & -- \\

Reproduced Baseline & VGG19 & PD-2 Ordinal & 65.6 & 48.9 \\

Reproduced Baseline & ViT & Cross Entropy & 68.59 & 69.42 \\

\textbf{Proposed} & 
\begin{tabular}[c]{@{}c@{}}ViT + CORAL +\\ Ensemble + TTA\end{tabular} 
& CORAL & \textbf{70.29} & \textbf{70.19} \\
\hline
\end{tabular}
}
\caption{Comparison with baseline supervised methods.}
\label{tab:baseline_comparison}
\end{table}

Overall, the proposed VL-OrdinalFormer consistently outperforms both CNN and transformer-based baselines by addressing three key limitations of prior approaches: lack of ordinal awareness, absence of clinical semantic guidance, and sensitivity to data imbalance. Unlike standard ViT classifiers trained with cross-entropy, our CORAL-based formulation embeds the natural ordering of disease severity directly into the learning process, reducing implausible grade transitions and improving discrimination between adjacent grades. Moreover, vision–language alignment via CLIP introduces clinically meaningful supervision by associating visual patterns with radiological concepts such as osteophytes, joint-space narrowing, and subchondral sclerosis. This semantic grounding is particularly beneficial for early and borderline cases (KL1–KL2), where visual differences are subtle and often misclassified by image-only models. Finally, ensemble inference combined with TTA and threshold optimization stabilizes predictions and mitigates overfitting, leading to balanced improvements across all KL grades. As a result, the proposed framework achieves superior accuracy, macro F1-score, and AUROC while maintaining ordinally consistent and clinically interpretable predictions.

\subsection{Ablation Studies}

To quantify the contribution of each component, we conduct a systematic ablation study by incrementally introducing the proposed design elements, i.e., CORAL-based ordinal regression, class-aware weighting emphasizing KL1–KL2, CLIP-based vision–language distillation, ensemble inference across folds, and TTA and threshold optimization.
\begin{table}[ht]
\centering
\caption{Ablation study of proposed model components.}
\label{tab:ablation}

\begin{tabular}{cccc}
\hline
\textbf{Configuration} & \textbf{\begin{tabular}[c]{@{}c@{}}Accuracy\\ (\%)\end{tabular}} & \textbf{\begin{tabular}[c]{@{}c@{}}F1(macro)\\ (\%)\end{tabular}} & \textbf{Notes} \\
\hline
\begin{tabular}[c]{@{}c@{}}ViT-CORAL\\ (no weights)\end{tabular} & 65.15 & 68.60 & Baseline ordinal \\
+ Class weighting & 66.4 & 63.6 & KL1/KL2 boosted \\
+ VLM Distillation & 67.2 & 65.0 & Semantic alignment \\
+ TTA + Tau tuned & 67.8 & 67.0 & Calibrated thresholds \\
\textbf{+ 5-fold Ensemble (final)} & \textbf{70.29} & \textbf{69.80} & State-of-the-Art \\
\hline
\end{tabular}
\end{table}
As summarized in Table~\ref{tab:ablation}, each component contributes to consistent performance gains. While ordinal regression establishes a strong foundation, the largest improvements arise from ensemble averaging and calibrated ordinal decoding. This confirms the complementary nature of the proposed components and their collective contribution to robust performance.

\subsection{Class-wise Improvement Analysis}

Table~\ref{tab:classwise} reports class-wise F1-scores for the reproduced VGG19 baseline, the reproduced plain ViT baseline trained with cross-entropy loss, and the proposed ViT–CORAL model. Moving from a CNN to a transformer backbone already yields consistent gains across most grades, particularly for KL2 and KL3, confirming the benefit of global context modeling in knee radiograph analysis. However, the plain ViT baseline still exhibits noticeable confusion in the early and mid-grade categories, where subtle structural changes dominate.
\begin{table}[!ht]
\centering
\caption{Class-wise F1-scores comparison (\%).}
\label{tab:classwise}
\begin{tabular}{llllll}
\hline
Method & KL0 & KL1 & KL2 & KL3 & KL4 \\ \hline
VGG19 baseline & 79.7 & 29.6 & 48.0 & 73.1 & 80.0 \\
ViT baseline & 76.03 & 35.09 & 64.47 & 77.46 & \textbf{86.79} \\
\textbf{Proposed} & \textbf{80.26} & \textbf{36.23} & \textbf{67.24} & \textbf{82.27} & 84.96 \\ \hline
\end{tabular}
\end{table}
In contrast, the proposed VL-OrdinalFormer achieves further and more consistent improvements across clinically challenging grades. Compared with the VGG19 baseline, it yields improvements of +6.6\% for KL1, +19.4\% for KL2, +9.2\% for KL3, and +4.9\% for KL4, directly targeting the primary source of diagnostic ambiguity in early and moderate KOA assessment. When compared against the plain ViT baseline, additional gains are observed for KL0 (+4.2\%), KL1 (+1.1\%), KL2 (+2.8\%), and KL3 (+4.8\%), demonstrating that ordinal modeling provides complementary benefits beyond transformer-based feature extraction alone.

Importantly, performance for KL0 remains stable or improves, indicating that enhanced sensitivity to mid-grade disease does not come at the expense of normal or advanced cases. Most remaining errors are confined within a $\pm1$ grade margin, reflecting ordinally consistent and clinically reasonable misclassifications. Figure~\ref{fig:confusion_matrix} further illustrates the improved prediction distribution across adjacent ordinal boundaries, with reduced long-range grade jumps compared to both CNN and softmax-based transformer baselines.

\subsection{Statistical Significance Analysis}

To assess the robustness of the observed improvements, we conduct paired statistical tests between:
(i) the reproduced VGG19 baseline and the final ensemble model, and (ii) the softmax-based ViT and CORAL-based ViT variants. For macro F1-score, paired t-tests yield:
\begin{equation}
p < 0.01,
\end{equation}

indicating statistically significant performance gains. In addition, bootstrap-based confidence intervals are computed for accuracy, confirming clear separation from baseline performance at the 95\% confidence level. These findings suggest that the improvements achieved by the proposed framework are unlikely to arise from random data partitioning and reflect genuine methodological advances.

\section{Interpretability and Analysis} \label{inter}

For medical AI systems to be clinically adopted, their predictions must be transparent and supported by interpretable evidence that clinicians can understand and validate. In the context of KOA severity assessment, interpretability is particularly important because diagnostic decisions rely on subtle radiographic cues. Accordingly, we integrate two complementary interpretability strategies: vision–language similarity heatmaps and feature attribution using Gradient-weighted Class Activation Mapping (Grad-CAM). Together, these techniques provide insight into the radiographic structures that drive model predictions and support clinical trust in automated KOA grading.

\subsection{CLIP-based Similarity Heatmaps}

The proposed VLM enables explicit alignment between visual features extracted from knee radiographs and radiologically grounded textual descriptions associated with KL grades. Visual embeddings from the image encoder and textual embeddings derived from KL-specific descriptions are projected into a shared latent space learned by the CLIP framework. These textual descriptors encode clinically meaningful severity cues, such as “no osteophytes and normal joint space” for KL0, “doubtful osteophytes” for KL1, “definite osteophytes with mild joint space narrowing” for KL2, “marked joint space narrowing” for KL3, and “bone-on-bone cartilage loss” for KL4.

Similarity-based attention heatmaps are generated by computing the correspondence between localized visual regions and the textual embedding of the predicted KOA grade. These heatmaps highlight anatomical regions that contribute most strongly to the model’s decision. For KL0 cases, activation patterns are diffuse and weak, reflecting the absence of radiographically evident pathology. In KL1 cases, attention becomes mildly concentrated around joint margins, consistent with early or doubtful osteophyte formation. KL2 cases exhibit clearer localization within the medial tibiofemoral compartment, indicating emerging joint space narrowing. For advanced stages (KL3 and KL4), strong and focused activation is observed along osteophyte boundaries and severely narrowed joint surfaces, capturing hallmark degenerative changes. These findings demonstrate that vision–language semantic alignment encourages anatomically meaningful reasoning and enhances the clinical interpretability of the model’s predictions, particularly for early and borderline disease stages.

\subsection{Grad-CAM Visualization of Radiographic Features}

To further analyze the internal decision-making behavior of the proposed ViT, we employ Grad-CAM with an attention-rollout strategy applied to the final prediction logits. This approach highlights image regions that contribute most strongly to the predicted  KL grade, enabling visual inspection of whether the model relies on clinically meaningful radiographic cues.

\begin{figure*}[!ht]
\centering
\includegraphics[width=1\textwidth]{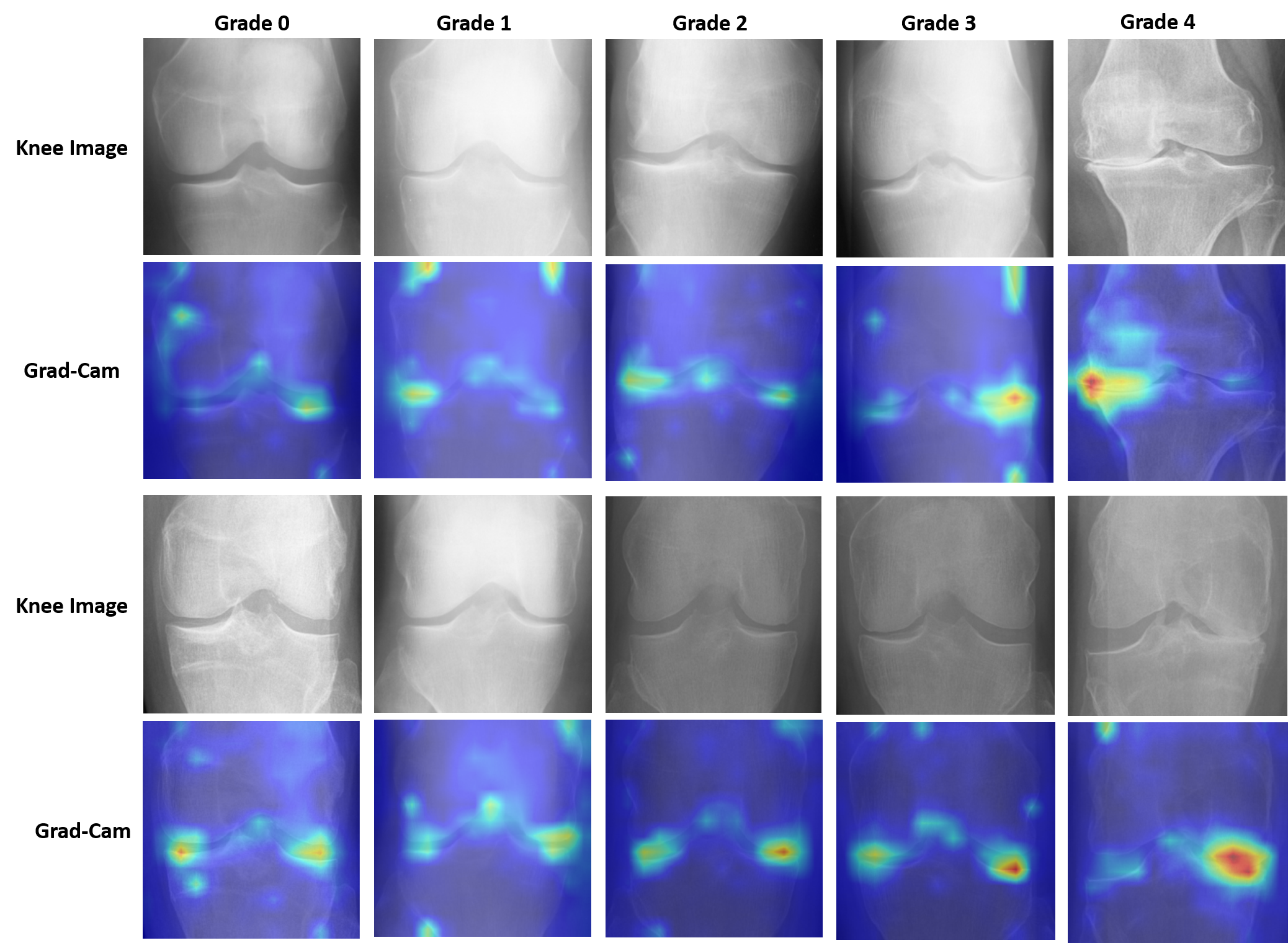}
\caption{Qualitative interpretability of the proposed ViT-L/16 CORAL model on representative OAI knee radiographs. The first and third row show the original knee images for KL grades 0-4, and the second and fourth row show the corresponding Grad-CAM. The highlighted regions concentrate around the tibiofemoral joint space and marginal osteophytes, indicating that the model focuses on clinically relevant structures when estimating KOA severity.}
\label{gradcam_vitl16}
\end{figure*}

For each KL grade, representative knee radiographs are selected and transformer attention maps are aggregated across layers and heads, projected back to the original $224 \times 224$ image space, and overlaid on the input radiographs. As illustrated in Fig.~\ref{gradcam_vitl16}, the resulting heatmaps consistently emphasize anatomically relevant structures, including the tibiofemoral joint space, marginal osteophytes, cortical irregularities, and subchondral bone contours, while largely ignoring background regions. For correctly classified KL3 and KL4 cases, activations are strongly concentrated along regions exhibiting pronounced joint space narrowing and bone remodeling, reflecting hallmark features of advanced KOA. In KL2 cases, Grad-CAM responses typically localize to regions of emerging medial joint space narrowing, which differentiates them from KL1 samples that display weaker and more diffuse activation patterns. Analysis of misclassified examples provides additional insight, when KL2 images are predicted as KL1, the corresponding heatmaps often emphasize superficial cartilage regions while failing to capture subtle medial collapse, suggesting opportunities for further refinement in fine-grained lesion representation.

Overall, these visual explanations demonstrate that the proposed ViT-L/16 CORAL model bases its predictions on clinically meaningful radiographic biomarkers rather than spurious background artifacts. The alignment between Grad-CAM activations and established diagnostic criteria supports the interpretability and clinical credibility of the proposed framework, enabling clinicians to qualitatively assess model reasoning alongside predicted KOA severity.




\section{Discussion}\label{discussion}

The objective of automated KOA assessment extends beyond achieving high predictive accuracy; it requires clinically faithful grading that reflects the progressive nature of the disease and aligns with expert diagnostic reasoning. The experimental results demonstrate that integrating vision–language modeling with ordinal learning substantially improves the discrimination of intermediate KL grades and enhances generalization across diverse patient populations. This section discusses the key factors underlying these improvements and their clinical implications.

\subsection{Impact of Vision-Language Alignment on Ordinal Classification}

KOA severity assessment is inherently ordinal, as radiographic abnormalities evolve gradually from KL0 to KL4. Conventional multi-class classifiers \cite{liu2020toward} typically treat grade labels as independent categories, which often results in ambiguous decision boundaries between adjacent grades, particularly KL1 and KL2. The proposed vision–language-guided framework addresses this limitation by introducing semantic supervision derived from clinically grounded textual descriptions. By associating visual patterns with interpretable descriptors such as joint space narrowing, osteophyte formation, and subchondral sclerosis, the model learns representations that more closely reflect expert-defined grading criteria. This semantic alignment effectively narrows the gap between image appearance and clinical interpretation, leading to reduced confusion between neighboring grades. The observed gains in macro recall and F1-score for KL1 and KL2 provide quantitative evidence that vision–language guidance enhances sensitivity to subtle radiographic biomarkers.

\subsection{Generalization and Robustness Across Variations}

Knee radiographs exhibit substantial variability due to differences in acquisition protocols, patient anatomy, exposure conditions, and positioning. Image-only models are prone to overfitting spurious textures or intensity patterns that may not correspond to true pathological changes. In contrast, vision–language-guided learning introduces semantic constraints that encourage the model to focus on anatomically and clinically meaningful structures. Moreover, the use of CORAL-based ordinal regression enforces smooth transitions between severity levels, avoiding abrupt decision boundaries and promoting stable predictions under distribution shifts. The consistent performance observed across five-fold cross-validation, along with balanced classification metrics, indicates that the proposed framework achieves improved robustness and generalization across heterogeneous data. While this study focused on KL-grade classification in KOA, the proposed ViT–CORAL pipeline is generalizable to other medical tasks with ordinal label structures. Examples include diabetic retinopathy staging, liver fibrosis grading, and cancer progression scores. Future work may explore adapting this approach to other modalities where ordinal reasoning is critical for clinical decision-making.

\subsection{Comparison with Radiologist Variability}

Previous clinical studies have reported notable inter-observer variability in KL grading, particularly for borderline cases such as KL1 and KL2, where Cohen’s kappa values often range from fair to moderate. In this context, perfectly deterministic classification is neither feasible nor clinically expected. Instead, the goal of automated systems should be to achieve performance comparable to expert consistency while maintaining transparency. The proposed approach demonstrates performance approaching reported levels of expert agreement, especially when ensemble inference, TTA, and threshold optimization are applied. Importantly, interpretability analyses show that the model consistently attends to anatomical regions and radiographic cues commonly used by radiologists, supporting the clinical plausibility of its predictions.

Overall, these findings suggest that vision–language-guided ordinal learning represents a meaningful advancement for automated KOA assessment. By jointly leveraging medical text semantics, ordinal disease structure, and radiographic imaging cues, the proposed framework reduces diagnostic ambiguity and enhances interpretability. This combination positions the method as a reliable and scalable tool for supporting radiological workflows and longitudinal KOA monitoring in real-world clinical settings.

\subsection{Implementation Details}

All experiments were implemented in PyTorch using the official ViT models provided by torchvision. Training and evaluation were conducted on an NVIDIA RTX 3090 GPU, with a system equipped with 64 GB of RAM. The proposed ordinal classifier is built upon a ViT-L/16 backbone pretrained on ImageNet-1K and adapted for five-grade KL severity prediction using a CORAL-based output formulation. An overview of the complete training and evaluation pipeline is provided in Algorithm~\ref{alg:train_vit_coral}, while key hyperparameters are summarized in Table~\ref{tab:hyperparams_vitl}. 
\begin{algorithm}[!ht]
\caption{Training pipeline of the proposed ViT-L/16 CORAL framework}
\label{alg:train_vit_coral}
\small
\begin{algorithmic}[1]

\Require Dataset $\mathcal{D}=\{(x_i,y_i)\}$, KL grades $y_i\!\in\!\{0..4\}$,
         number of folds $K$, batch size $B$, learning rate $\eta$,
         class weights $w_{\text{cls}}$, distillation weight $\lambda$
\Ensure Trained fold models $\{M_k\}_{k=1}^K$

\State \textbf{Pre-processing}
\State Resize all images to $224{\times}224$ and normalize (ImageNet statistics).
\State Define clinical text prompts $t_c$ and obtain CLIP text embeddings $z_c$.

\Statex
\State \textbf{Cross-validation and training}
\State Stratify-split $\mathcal{D}$ into $K$ patient-level folds.
\For{$k=1$ to $K$}
    \State Construct $\mathcal{D}^{train}_k$, $\mathcal{D}^{val}_k$, $\mathcal{D}^{test}_k$.
    \State Load ViT-L/16 backbone (ImageNet weights) and replace classifier with CORAL head.
    \State Initialize optimizer (e.g., AdamW) and learning-rate scheduler.

    \For{each epoch}
        \For{each mini-batch $\{(x_b,y_b)\}_{b=1}^B \subset \mathcal{D}^{train}_k$}
            \State Apply data augmentation to $x_b$.
            \State Extract CLS feature $h_b = \mathrm{ViT\_L16}(x_b)$.
            \State Compute CORAL logits $o_b = f_{\text{CORAL}}(h_b)$.
            \State Convert $y_b$ to ordinal labels $y_b^{ord}$.
            \State $\mathcal{L}_{\text{coral}} =
                   \mathrm{BCEWL}(o_b, y_b^{ord}, w_{\text{cls}})$.

            \If{VLM distillation is enabled}
                \State Compute cosine scores $s_{b,c}$ and teacher distribution $p^{\text{VLM}}_b$.
                \State Derive student distribution $p^{\text{img}}_b$ from $o_b$.
                \State $\mathcal{L} =
                       \mathcal{L}_{\text{coral}} +
                       \lambda\,\mathrm{KL}\!\left(p^{\text{VLM}}_b \,\|\, p^{\text{img}}_b\right)$.
            \Else
                \State $\mathcal{L} = \mathcal{L}_{\text{coral}}$.
            \EndIf

            \State Update network parameters $\theta$ using $\nabla_\theta \mathcal{L}$.
        \EndFor

        \State Evaluate on $\mathcal{D}^{val}_k$ and monitor MAE / F1(macro) for early stopping.
    \EndFor

    \State Save best-performing model as $M_k$.
\EndFor

\end{algorithmic}
\end{algorithm}
The dataset is organized into separate training, validation, and test directories, each containing class-specific subfolders corresponding to KL0–KL4. During model development, the original training and validation sets are merged, and stratified five-fold cross-validation is performed at the image level to preserve the KL grade distribution in each split. For each fold, four subsets are used for training and one subset is reserved for validation. All knee radiographs are resized to a spatial resolution of $224\times224$ pixels. During training, data augmentation includes random resized cropping, horizontal flipping, and small-angle rotations to improve robustness to variations in acquisition conditions and patient positioning. Validation and test images undergo center cropping and normalization only. All inputs are normalized using the ImageNet RGB mean and standard deviation.

The ViT-L/16 backbone is initialized with ImageNet-1K pretrained weights. The original softmax classification head is removed and replaced with a linear layer producing $(K-1)=4$ outputs, where $K=5$ corresponds to the number of KL grades. These outputs represent CORAL logits associated with ordered decision thresholds. For a ground-truth label $y \in {0,1,2,3,4}$, CORAL targets are defined as binary indicators $\mathds{1}[y > k] $ for $k = 0,\dots,3$. Training minimizes a weighted binary cross-entropy loss across all thresholds. To address threshold-specific class imbalance, a pos\_weight term is computed from the empirical label distribution, and additional per-sample class weighting is applied to emphasize the clinically challenging KL1 and KL2 categories.

\begin{table}[!ht]
\centering
\caption{Training and model hyperparameters for the ViT-L/16 + CORAL KL grading experiments.}
\label{tab:hyperparams_vitl}
\begin{tabularx}{\textwidth}{lX}
\toprule
\textbf{Component} & \textbf{Setting} \\
\midrule
Backbone architecture & ViT-L/16 (ImageNet-1K pre-trained) \\
Output formulation & CORAL ordinal regression, $K-1=4$ logits \\
Number of classes & $K=5$ (KL0-KL4) \\
Input resolution & $224\times224$ pixels, RGB \\
Batch size & 8 images per GPU \\
Optimizer & AdamW \\
Initial learning rate & $3\times10^{-5}$ \\
Weight decay & 0.05 \\
LR scheduler & Cosine annealing ($T_{\max}=80$ epochs) \\
Maximum epochs per fold & 80 \\
Early stopping patience & 10 epochs (validation accuracy) \\
Loss function & BCE-with-logits over thresholds (CORAL) \\
Threshold \texttt{pos\_weight} & Data-driven, $w_k = N_{\text{neg}} / N_{\text{pos}}$ per threshold \\
Per-class sample weights & $[1.0, 1.5, 1.5, 1.0, 1.0]$ for KL0-KL4 \\
Number of folds & 5-fold stratified cross-validation \\
Train transforms & Resize to $224$, random resized crop (0.8-1.0), horizontal flip, random rotation ($\pm10^\circ$), normalization \\
Val/test transforms & Resize to $224$, center crop, normalization \\
Test-time augmentation (TTA) & Identity, horizontal flip, $+10^\circ$ and $-10^\circ$ rotations \\
Inference decoding & Count of thresholds with $\sigma(\ell_k)\ge\tau$ \\
Tau tuning range & $\tau \in [0.30, 0.70]$ (step 0.01), selected by macro-F1 on validation \\
Random seed & 42 (Python, NumPy, PyTorch) \\
\bottomrule
\end{tabularx}
\end{table}

For each fold, training proceeds until either the maximum number of epochs is reached or early stopping is triggered when validation accuracy fails to improve for ten consecutive epochs. The checkpoint corresponding to the best validation accuracy is saved for each fold. After completing all five folds, a second-stage threshold calibration is performed. Specifically, checkpoints from all folds are reloaded, and test-time augmented logits are collected on each fold’s validation set. These logits are concatenated across folds, and a global decision threshold $\tau$ is selected from the interval $[0.30, 0.70]$ with a step size of $0.01$ by maximizing the macro F1-score. This single calibrated threshold is then fixed for all subsequent evaluations. During testing, two evaluation settings are considered. First, the best-performing single-fold model is evaluated using TTA and the tuned threshold $\tau$. For each test image, four augmented variants (original image, horizontal flip, and rotations of $+10^\circ$ and $-10^\circ$) are generated. Logits are averaged across augmentations, passed through a sigmoid function, and decoded into an ordinal prediction by counting the number of thresholds exceeding $\tau$. Performance metrics and confusion matrices are computed for this single-model setting. 

Second, a five-model ensemble is evaluated to exploit complementary information learned across different training splits. For each test image, test-time augmented logits are computed independently for each fold-specific model, averaged across augmentations, and then averaged across models before applying the same threshold-based decoding rule. Ensemble predictions are used to compute accuracy, macro- and weighted-averaged precision, recall, F1-score, specificity, and macro AUROC. 

As shown in Table \ref{tab:hyperparams_vitl} the learning rate of $3\times10^{-5}$ and weight decay of $0.05$ were selected based on established practices in ViT fine-tuning, where large-capacity models benefit from conservative learning rates and moderate regularization to ensure stable convergence. The AdamW optimizer was chosen due to its decoupled weight decay formulation and demonstrated robustness in transformer-based architectures. A cosine annealing learning rate schedule is employed to provide a smooth decay over training epochs, which empirically improves convergence stability on the OAI dataset. A batch size of 8 represents a practical trade-off between GPU memory constraints and reliable gradient estimation for the ViT-L/16 backbone. Early stopping with a patience of ten epochs mitigates overfitting on the relatively small validation folds while avoiding unnecessarily prolonged training. To address class imbalance and the clinical importance of intermediate disease stages, per-class sample weights of $[1.0, 1.5, 1.5, 1.0, 1.0]$ are applied, modestly emphasizing KL1 and KL2 without introducing instability. Finally, the global threshold ($\tau$) tuning strategy is designed to optimize the macro F1-score on validation data, reflecting balanced performance across all KL grades. This post-hoc calibration improves ordinal boundary placement while preserving a simple and interpretable decision rule at deployment time, which is important for clinical adoption.

\section{Conclusion and Future Work}\label{con}

This study introduced a vision–language-guided ordinal learning framework for fully automated KOA severity assessment from radiographic images. By integrating a ViT-L/16 backbone with CORAL-based ordinal regression and CLIP-driven semantic supervision, the proposed method achieves consistent improvements over existing baselines. In particular, it outperforms the widely used VGG19 and ViT frameworks in terms of accuracy and macro-level classification metrics, with notable gains for the clinically challenging KL1 and KL2 categories, where subtle morphological changes often lead to inter-observer disagreement. Complementary interpretability analyses using Grad-CAM and CLIP similarity maps further demonstrate that the model attends to clinically relevant structures, including joint space width, marginal osteophytes, and subchondral bone characteristics. The use of ensemble inference with TTA contributes to improved prediction stability and robustness against anatomical variability and acquisition-related differences, strengthening the suitability of the approach for real-world clinical settings. Together, these findings indicate that combining ordinal constraints with vision–language representations provides an effective and interpretable paradigm for automated KOA grading. Future work will explore several extensions of the proposed framework. First, incorporating bilateral and multi-view radiographs may enable more comprehensive modeling of joint asymmetry and tibiofemoral alignment, which are important indicators of KOA progression. Second, aligning model outputs with structured radiology report terminology could support natural-language explanations and facilitate integration into existing clinical workflows. Finally, evaluation on large-scale, multi-center datasets will be pursued to assess generalizability across diverse patient populations, imaging protocols, and clinical environments. These directions aim to further advance automated KOA assessment toward reliable and scalable real-world deployment.



\section*{CRediT authorship contribution statement}
\textbf{Zahid Ullah:} Conceptualization, Methodology, Software, Formal analysis, Investigation, Writing - original draft, Writing - review \& editing.  \textbf{Jihie Kim:} Formal analysis, Investigation, Supervision, Project administration.  

\section*{\textbf{Declaration of Competing Interests}} The authors declare that they have no known competing financial interests or personal relationships that could have appeared to influence the work reported in this paper.


\section*{Acknowledgements}
This research was supported by the MSIT(Ministry of Science and ICT), Korea, under the ITRC(Information Technology Research Center) support program(IITP-2026-RS-2020-II201789), and the Artificial Intelligence Convergence Innovation Human Resources Development(IITP-2026-RS-2023-00254592) supervised by the IITP(Institute for Information \& Communications Technology Planning \& Evaluation).






\bibliographystyle{elsarticle-num-names}
\bibliography{sample.bib}







\end{document}